# KartalOl: Transfer learning using deep neural network for iris segmentation and localization: New dataset for iris segmentation

Jalil Nourmohammadi Khiarak [*1a], Samaneh Salehi Nasab[b], Farhang Jaryani[c], Seyed Naeim Moafinejad[d], Rana Pourmohamad[e], Yasin Amini[f], Morteza Noshad[g]

[*] Institute of Control and Computation Engineering, Warsaw University of Technology, Warsaw, Poland
[b] School of Engineering, Lorestan University, Aleshtar Campus, Aleshtar, Iran.
[c] Faculty of Engineering, Arak University, Arak, Iran.
[d] Department of Physics, Shahid Beheshti University, Tehran, Iran.
[e] School of Computing, Informatics and Decision Systems Engineering, Arizona State University, Arizona, USA.
[f] Faculty of Electrical and Computer Engineering, University of Kharazmi, Iran.
[g] Senior Research Scientist, Stanford University, Stanford, USA.

## ABSTRACT

Iris segmentation and localization in unconstrained environments is challenging due to long distances, illumination variations, limited user cooperation, and moving subjects. To address this problem, we present a U-Net with a pre-trained MobileNetV2 deep neural network method. We employ the pre-trained weights given with MobileNetV2 for the ImageNet dataset and fine-tune it on the iris recognition and localization domain. Further, we have introduced a new dataset, called KartalOl, to better evaluate detectors in iris recognition scenarios. To provide domain adaptation, we fine-tune the MobileNetV2 model on the provided data for NIR-ISL 2021 from the CASIA-Iris-Asia, CASIA-Iris-M1, and CASIA-Iris-Africa and our dataset. We also augment the data by performing left-right flips, rotation, zoom, and brightness. We chose the binarization threshold for the binary masks by iterating over the images in the provided dataset. The proposed method is tested and trained in CASIA-Iris-Asia, CASIA-Iris-M1, CASIA-Iris-Africa, along the KartalOl dataset. The experimental results highlight that our method surpasses state-of-the-art methods on mobile-based benchmarks. The codes and evaluation results are publicly available at https://github.com/Jalilnkh/KartalOl-NIR-ISL2021031301.

## 1. Introduction

The advent of technology has impacted biometrics recognition techniques such as fingerprint, face recognition, iris recognition, and periocular recognition. Among these biometric identification technologies, iris-based methods are highlighted as the most reliable and accurate technology [1, 2], therefore, it has attracted the biometric community [3].

Each iris recognition system typically consists following sub-process: iris image acquisition, preprocessing, iris segmentation, iris feature extraction, and iris matching verification or identification [4]. Iris segmentation and localization determine iris region pixels in an image; this applies for feature extraction and matching [5]. It is a critical step for improving the accuracy of iris recognition. [6] as by determining the proper iris region, we can obtain worthy information from the iris image and additional advance the efficiency of the iris recognition system [4].

The majority of acquired iris images necessarily contain noise, such as occlusions caused by eyelids or eyelashes, specular reflections, off-angle, and blur. To make full use of these noisy iris images, effective and reliable iris segmentation and localization have been recognized as the first and most important problem still facing the biometric community, affecting all downstream activities from normalization to recognition [7].

Furthermore, many of the past efforts have been undertaken mostly in controlled environments. They are not very robust, and their primary goal is to segment the noise-free mask while ignoring the iris boundary parameterization. This research tries to address this issue by developing a new model.

The main contributions of this work can be summarized as follows:

1- Using a modified deep neural network, we propose a comprehensive model for iris segmentation and localization.

2- Introducing a new dataset for iris segmentation and localization called KartalOl.

3- Proposing a method that has efficiently performed NIR iris images in unconstrained, specular reflections, off-angle, blur, and noise environments. It also improved the state-of-the-art in mobile-based iris images.

The following is a breakdown of the paper's structure: Sect. 2 examines some previous research as well as the rationale for the proposed effort. The proposed iris segmentation and localization are implemented in Section 3. The proposed method's experimental findings are shown in Section 5. Finally, in the last section, the conclusions are presented.

## 2. Literature review

Li and colleagues [8] proposed a deep-learning-based iris segmentation technique. To mark and identify eye location, the authors first create a quicker R-CNN supplemented with six layers. The Gaussian mixture model is then used to find the pupil. They then use a set of five crucial boundary points to identify the circular iris inner boundary. The boundary-points election procedure is then used to determine the iris outer boundary-points, which are then utilized to localize the iris outer boundary. On a dataset with eyelashes that are not thick or bunchy, this method performed better. The detection of the iris inner boundary may be the scheme's biggest flaw. It may function poorly in dimly lit areas (e.g., bunches of dark eyelashes, congested eyebrows, and hair strips).

---

[1] Corresponding author. Tel.: +48-510-608-921; e-mail: Jalil.Nourmohammadi@elka.pw.edu.pl

Han et al. [9] focused on iris localization, proposing a practical iris localization approach for noisy iris pictures. The two steps in their proposed iris localization method are pupil border localization and iris boundary localization. To localize a pupil region, an efficient block-based minimum energy detection method is used, with specular reflection removal as a preprocessing step. The NICE.II dataset was used in this study's experiments. The dataset includes a variety of noisy photos captured in the real world.

Gad et al. developed an iris-based recognition technique as a unimodal biometric using multi-biometric situations. During the segmentation phase, a novel method based on the masking approach was developed to find the iris. Two novel techniques, delta-mean and multi-algorithm-mean, were developed to extract iris feature vectors. CASIA v. 1, CASIA v. 4-Interval, UBIRIS v. 1, and SDUMLA-HMT were used to test the suggested system. The proposed solution for authentication concerns performs satisfactorily, according to the results [10].

Another study is proposed to extract region-based information from non-cooperative iris pictures using an adjustable filter bank [11]. The proposed approach is based on a 14th order half-band polynomial. Data was trained in the CASIAv3, UBIRISv1, and IITD dataset, and filter coefficients were collected from the polynomial domain rather than the z-domain. Texture features were extracted from annular iris templates that were suitably localized using an integrodifferential operator using the tunable filter bank. The original iris template is initially separated into six equispaced sub-templates.

Singh et al. [12] offer an iris identification technique based on feature extraction using the Integer Wavelet Transform (IWT). The variation (RTV) model is used in conjunction with other models. Four-level IWT is used to normalize and deconstruct the segmented iris area. To speed up iris segmentation, simple filtering, Hough transform, and edge detection are used to approximate the location of the iris. The input image is transformed into 256 sub-bands using a four-level Integer Wavelet Transform, of which only 192 lower sub-bands are considered. High-frequency sub-bands are ignored since they introduce noise to the system and reduce its precision. Energy is recovered from each of the 192 sub-bands, resulting in a 192-bit binary code. The energy of each sub-band is compared to a pre-computed individual tailored threshold value to generate a unique iris code.

## 3. Database

We have collected a database for visible spectrum iris photos and more importantly for using smartphone camera, namely KartalOl dataset.

The fundamental issue with all iris images is noise, occlusion, blur, specular reflection, and off-angle. As a result, the biometric community's most pressing unsolved difficulty affects all tasks ranging from iris normalization to recognition using these noisy iris images. (Segmentation and Localization of the NIR Iris Challenge in Non-cooperative Environments) Publicly available datasets for human iris photographs serve an essential role in iris recognition research. The accessible datasets share several characteristics, such as near-infrared imaging, and adhere to John Daugman's [13] requirements. We now have additional iris picture datasets [14, 15] thanks to advances in mobile computing and deep learning (DL) in biometric applications. Finding a suitable dataset to complete one's research activities can be difficult for researchers who are new to this field. Since most dataset providers only allow noncommercial research to be done with their datasets.

Furthermore, authors usually strictly adhere to the access terms and require a signature from the researcher or a legal representative of the research institution. This imposes additional limitations on the attractiveness of particular datasets among academics. This study presents a new dataset for visible spectrum iris research that includes iris photos taken with two different smartphones. Two of the latest and most popular phones have been added to the KartalOl dataset: the iPhone X and iPhone 11. The photos were taken with a smartphone's rear camera.

The images in this dataset were taken by volunteers mainly from Central and South Asian countries. This collection contains iris photos from Central and South Asia, which are useful for evaluating the effectiveness of iris recognition algorithms across ethnic groups. The KartalOl dataset can be used to assess the robustness of several steps in an iris identification pipeline, such as segmentation and feature extraction, and thus reveal dependencies related to iris color and ethnicity. Another crucial consideration is the type of smartphone used to collect data for this dataset. This dataset was created using the most recent iPhones, the iPhone X and iPhone 11, to compare their iris recognition capabilities. In mixed illumination, the KartalOl dataset is collected.

Fig. 1 shows a selection of images from the KartalOl collection. This data was gathered from people from all across the world. It is divided into two distinct categories (the morning group and the evening group).

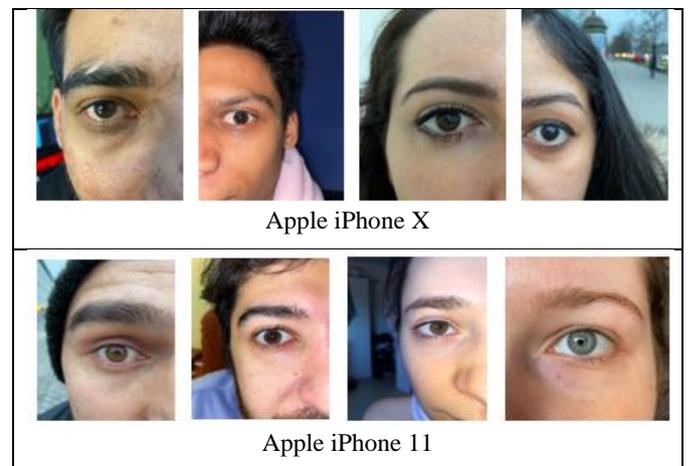

Fig 1: Sample images from the KartalOl dataset acquired using two different phones (Apple iPhone X and Apple iPhone 11).

Fig 2. highlights the gender and iris color distributions. The subset of data under the gray/green label indicates the group bearing the colors of iris such as light brown, dark brown.

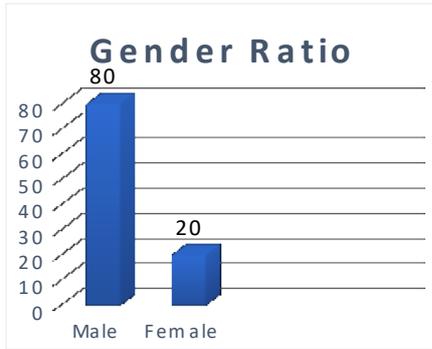

Fig 2: Color and gender distribution of samples in dataset

**Mask for our dataset**

Intel develops a new program, namely Computer Vision Annotation Tool (CVAT), to speed up the annotation process. Computer Vision Annotation Tool (CVAT) is applied to carry out the Iris Mask Segmentation. CVAT is a free online tool with interactive capability for image and video annotation, especially computer vision. We used the CVAT tool in order to annotate a staggering number of objects with different properties. Many decisions regarding UX and UI design are made according to the feedback received from the data annotation experts. There are various methods to conduct data annotation, but applying a proper specific tool will increase the processing speed. Advantages of CVAT include:

1. A web-based application means the user does not need to go through application installation. A user only requires to open the browser and create a task or do the data annotation.
2. It is simple to implement CVAT. It is possible to deploy it on a local network by using Docker.
3. It makes annotation easy by automatic option. For instance, users can employ interpolation between keyframes.
4. 
5. It is developed professionally. This tool was developed by an expert team of annotation and algorithmic.
6. CVAT is an open-source program under the MIT license.

Fig. 3 represent the annotation process in detail.

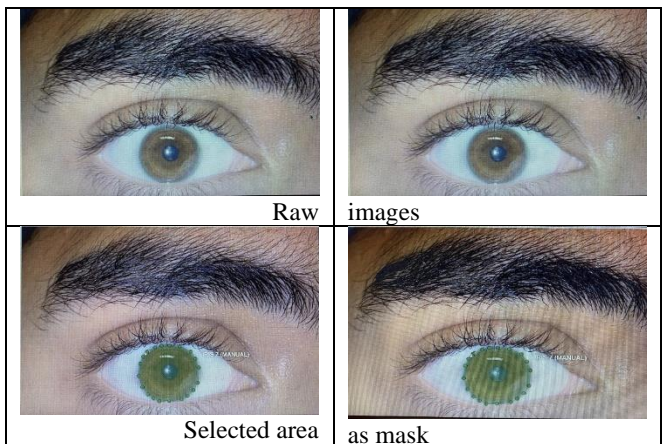

Fig. 3: Mask annotation for iris recognition and segmentation using CVAT tools.

### 4.2. Iris localization

To get inner and outer boundaries in iris images, we proposed a deep neural network-based localization method using pretrained

## 4. Proposed method

In this section, the fundamental steps of our proposed method are explained. Our focus is on two crucial issues for iris localization and iris segmentation. Firstly, we explain iris segmentation and illustrate the method that we have proposed for this purpose. Following that, we describe our iris localization method, and finally, we elaborate on a simple iris recognition method to evaluate our collected dataset.

### 4.1. Iris Segmentation

We aim to study UNet segmentation [16] and utilize it in iris image segmentation while using pre-trained MobileNetV2 [17] as the encoder for the UNet architecture. The purpose of integrating the pre-trained MobileNetV2 with the UNet is to obtain a sufficient network architecture. Training MobileNetV2 is carried out on the ImageNet dataset, among the most frequently utilized datasets. The pipeline of iris segmentation is shown in Fig 4. As it is shown, training step uses iris images and masks as inputs and for main body of proposed algorithm U-Net alone with MobileNetV2 is used then at output we have produced mask. In test phase, iris images are supply as input and masks are obtained using the trained model.

A down-sampling part and an up-sampling part are two parts of the U-net model, in which high-level features are fused with low-level features via a shortcut between two parts, improving the capability of segmenting image details. The feature extraction layers of pre-training MobileNetV2 substitute the down sampling part. The presence of four inverted residual blocks and five deconvolution layers in the up-sampling part confirms the identical form of the input and output dimensions of Mobile-Unet. Fig 6 depicts the architecture of Mobile-Unet. The detail of operations for each layer is listed in Table 1, in which ConvTranspose# and Inverted Residual# denote the deconvolution layer and inverted residual blocks, respectively; s, t, and k represent stride, expansion factors, and convolution kernels, respectively. Besides, the input of the model is the original fabric image; the output is the mask.

Table 1: Definition and operation of each layer of MobileNetV2-Unet.

| Layers | Input | Output | Output size | k | s | T |
|---|---|---|---|---|---|---|
| Input image | - | - | (3,224,224) | - | - | - |
| D1 | Input image | X1 | (16,128,128) | - | - | - |
| D2 | X1 | X2 | (24,64,64) | - | - | - |
| … | … | … | … | … | … | … |
| Inverted Residual4 | L7+X1 | L8 | (16,128,128) | - | 6 | 1 |
| ConvTranspose5 | L8 | Output | (1,224,224) | 4 | - | 2 |

MobileNetV2 alone with UNet. The pipeline of the proposed method is shown in Fig 4. As shown in Fig 4, we first use the pre-trained iris segmentation method to find the iris area, then we find the center of the iris and crop it. After that, our deep neural network would receive specific area of eye images as input. It means our DNN model is trained images which has an iris area mostly. According to our experience, DNN works well when we remove extra information and focus on the desired object in an image. As a result, in the evaluation part, we start from the original image, and after finding iris localization, we map it to the original image.

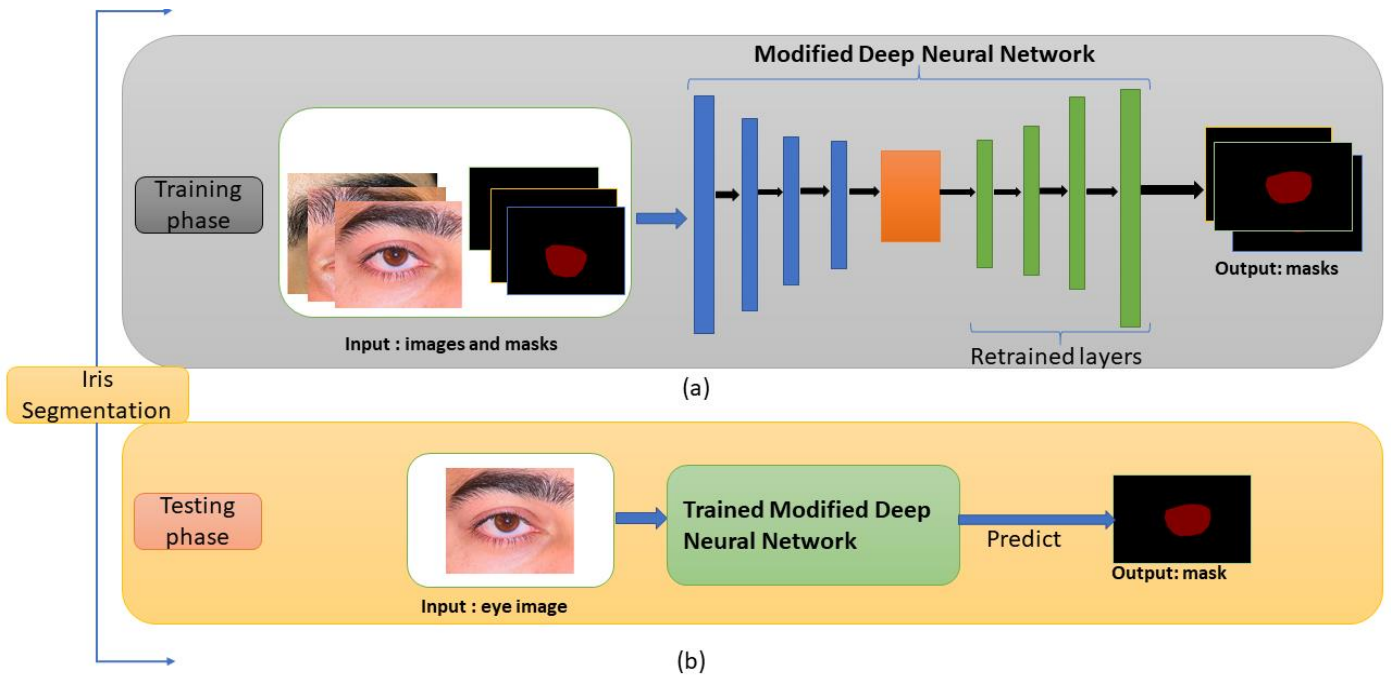

Fig 4: The pipeline of iris segmentation with two phases training and test phase, a) training phase: in this step mask and images are applied and mask as output is obtained, b) testing phase: in this step an eye image is represented as input and a mask is obtained as output.

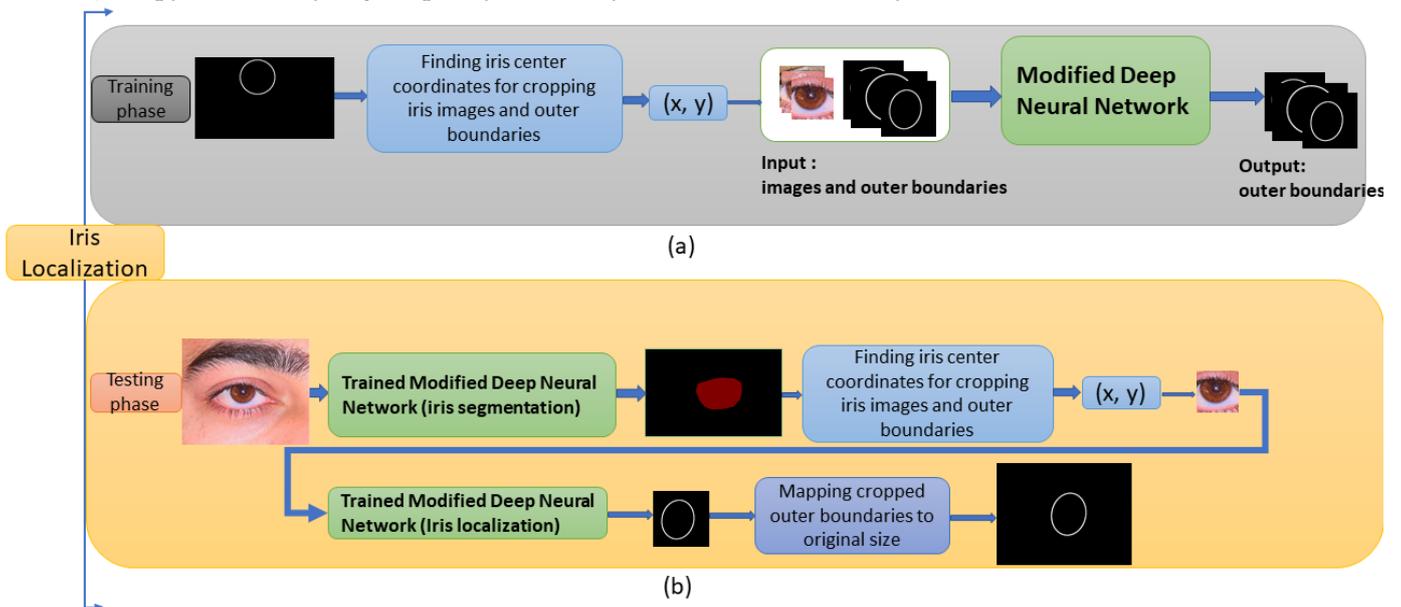

Fig 5: The pipeline of iris localization with two phases training and test phase, a) training phase: in this step segmentation of iris is applied using iris segmentation step then the iris region is cropped and mask and iris images are considered as input and mask as output is obtained, b) testing phase: in this step an eye image is represented as input then iris segmentation method is applied to find iris region and then using trained iris localization method a mask is obtained as output.

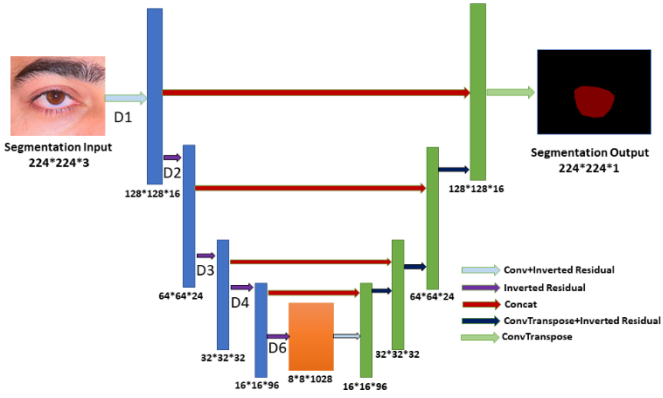

Fig 6: The architecture of Mobile-Unet.

## 5. Experimental results and discussion

### 5.1. Experimental setup

Four datasets have been used to access iris images, including CASIA-Iris-ASIA [7], CASIA-Iris-Mobile [7], CASIA-Iris-Africa [7], and our collected dataset called KartalOl.

When looking at a person, they will almost always have both their left and right eye images. This study utilized a dataset from KartalOl that contains a total of 300 photographs taken by 65 people over the course of two separate sessions. These pictures include a variety of noise components, which are intended to simulate less restrictive acquisition settings when capturing photographs. Many of the images in the CASIA-Iris-ASIA collection are taken from the participants' left and right eyes, which is not surprising. Through the use of uniformity and cross-validation, we determined that each subject in CASIA-Iris-Africa is applied if it has an equal number of iris photographs or more than five images on either the left or right eye branch on each side of the iris. CASIA-Iris-Africa includes a total of five pictures for each subject. More importantly, in cases when both the left and right eye branches meet the criterion, we select the left eye branch to represent the subject. After doing a requirement analysis, it was discovered that only 50 individuals from CASIA-Iris-Africa fulfilled the study's qualifying criteria. Each participant in the studies is referred to as a class throughout the experiment.

In this experiment, five iris images are randomly selected from each of the mentioned subjects. One image is used for the testing sample, while the other four were used for training samples. It is not easy to obtain enough images per subject for training and testing in most real applications. Therefore, this setting was made to simulate the practical application. The following section shows the all-recognition rate, which is the average result of five cross-verification experiments. There are 300 iris images from 100 subjects in the CASIA-Iris-ASIA dataset. Eyeglasses and specular reflections are the primary sources of intra-class variations. These variations are applied to examine images in a harsh environment. It is also used to calculate similarities. A cosine similarity metric is utilized for similarity calculation. As highlighted by previous research, Cosine similarity metrics are a powerful method for image recognition problems [18]. It is used to classify test samples into the same class as the nearest sample from the training set. In general, it is applied to measure the distance when the magnitude of the vectors is not essential. Usually, this occurs during working with image data characterized by some features. We can presume that when a feature (e.g., length of edges) happens more frequently in image 1 than in image 2, image 1 is more related to the length of edges. Then, the length of edges probably happened frequently in image 1 as it was way longer than image 2. Cosine similarity corrects this.

The appropriate example of using this metric would be image data. Although, cosine similarity might be applied for other cases where instances properties cause larger weights without any distinct difference.

### 5.2. experimental protocol

Based on four distinct types of training data I CASIA-Iris-Africa, ii) CASIA-Iris-Africa and KartalOl dataset), we train the algorithm in this article, taking into consideration the various components of the datasets as well as application situations. During that, following the training step of the algorithm, we were able to get a model that was tailored to the particular target. Finally, during the testing step, we conducted a regular within-dataset assessment of the models that had been trained. The results of the testing include binary iris segmentation masks matching to the test pictures, with non-zero valued pixels indicating the foreground (excellent iris texture or border) area and zero-valued pixels representing everything else in the images.

**Iris Segmentation**; When it came to segmenting and localizing the iris, each method was assessed for its efficacy. The following assessment metrics were generated based on the binary segmentation predictions and associated ground truths for all techniques, and they are as follows:

Iris segmentation: To evaluate the performance of iris segmentation, we used the two criteria provided by the NICE-I competition. The most significant statistic, E1, is the average proportion of matching disagreeing pixels over all pictures, which may be determined by using a pixel-wise XOR operator across the anticipated and ground truth iris masks.

$$E1 = \frac{1}{n*h*w}\sum_i\sum_j M(i,j) \otimes G(i,j) \quad (1)$$

Where i and j are the pixel coordinates in the predicted iris mask M and the ground truth iris mask G, respectively, and h and w are the width and height of the testing image, and n is the number of testing photos.

The second metric, E2, is used to compensate for the difference between the prior probability of "iris" and "non-iris" pixels in the images' prior probabilities. The average of the false positives (FP) and false negatives (FN) rates, to put it another way, is the rate of false positives.

$$E2 = \frac{1}{n*2}\sum_i (f_p + f_n) \quad (2)$$

FP shows the balance of background pixels wrongly retrieved as iris pixels, FN means the proportion of iris pixels inaccurately retrieved as background pixels, and n is the number of testing images.

E1 and E2 are defined in [0; 1], where the lower and higher values serve sequentially better and worse iris segmentation results.

**Iris localization:** In this work, we have utilized two common metrics for iris localization performance assessment, which are the Dice index and the Hausdorff distance, to evaluate iris localization performance. The Dice index and the Hausdorff distance are both used to evaluate iris localization performance.

The dice index is the calculating area of overlap in two images. It means each recognized utter boundary or inner boundary image is compared with grand truth.

Hausdorff distance is the longest of all the distances from any points in one set (utter boundary or inner boundary in iris images) to the closest points in the other utter boundary or inner boundary of iris images.

### 5.3. Quantitative Evaluation

The proposed method is assessed using a variety of metrics on a variety of testing sets based on the experimental setting. Multiple computed metric scores and associated rankings are ultimately combined to compare the proposed overall performance. We show the prediction results for iris segmentation and localization, then synthesize the findings to analyze the suggested method's performance. Table 3 presents a summary of the overall assessment findings obtained by the suggested approach across all testing sets.

**Result of training the proposed method:** We used 3530 images to train the model. The dataset was divided into three sections: training, validation, and test. We used 3178 items of training data, 176 pieces of validation data, and 176 pieces of testing data. Hyper-parameter optimization, which involves selecting the appropriate collection of parameters to provide well-fit weights and biases for the deep learning U-net model, is a critical step for training an effective ML model. Compiler (Adam), Learning rate (0.0001), and accuracy metric were all used. Negative dice coefficient [9] (to be maximized). Loss metric based on dice coefficients (to be minimized)

The model complexity is another parameter that needs to be tweaked (for instance, the training complexity forced by the size of the U-net). As a result, training on deep U-nets (depth=4) and shallow U-nets (depth=3) is required. For "small data challenges," the model with fewer parameters is usually the best option. The loss curves at the end of model training illustrate how effective the training procedure was. Fig 7 and Fig 8 below demonstrate the loss curves for our implementation.

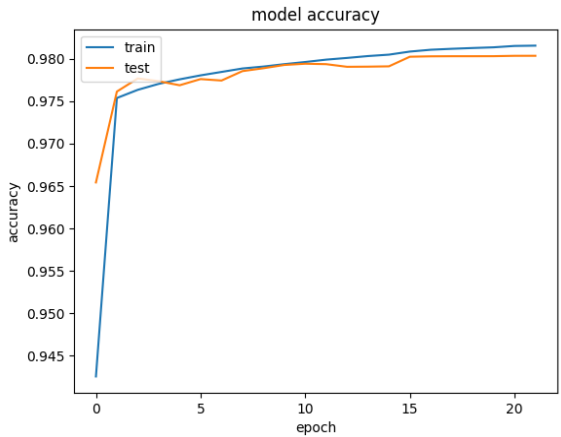

Fig 7: Accuracy of our proposed method on the dataset

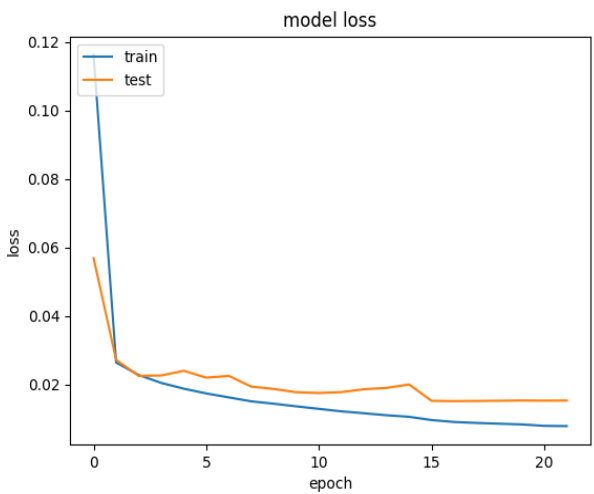

Fig 8: Loss value of our proposed method on the dataset

The above model was able to predict with an accuracy of 98% on the test set. There are just two publicly available open-source results and code of the competition, namely EyeCool and Lao Yang Sprint Team.

**EyeCool:** A newly developed U-Net model is used to segment the iris mask, as well as the iris's inner and outer boundaries, all at the same time. There are two types of enhancements. The EfficientNet-B5 is utilized for feature extraction, and a similar technique is suggested to improve the encoder's ability to take iris borders into consideration. Essentially, the goal of this technique is to insert many side-out layers after the decoder convolutional layers in order to forecast the iris boundary heatmaps, and then utilize these heatmaps as the weight in order to force the decoder to consider the iris borders [7].

**Lao Yang Sprint Team:** ResNet-34 as a backbone, but with a U-Net-based design. As an extra job, this group examines iris localization in order to tackle the iris segmentation problem by segmenting the mask within the iris border. U-Net in two separate models for iris segmentation and localization solves the poor performance of the multitasking model for these two tasks where the only variation between the segmentation and localization networks is the final head layer. Two parallel convolution layers are used by the localization head to produce the inner and outer border masks, whereas just one convolution layer is used by the segmentation head to produce the iris mask. The Lao Yang Sprint Team used dice loss with binary cross-entropy loss as part of this technique in addition to significant data augmentation [7].

Results of Iris Segmentation and localization. It is shown in Table 1 that the metric rankings on all testing sets are totaled to represent the overall segmentation and localization performance on various datasets, which are decided based on the respective metric score. The total Dice index for the inner or outer iris margin was calculated by averaging similar Dice indices across all test pictures. Lastly, the challenge utilized the mean value of the total mDice iris Dice indexes for ranking. Similar to the Dice index, the challenge calculated the mean normalized Hausdorff distance for the iris' inner and outer margins. Finally, ranking relied on the average of two normalized Hausdorff distances, referred to as mHdis.

As can be seen from Table 2, different methods demonstrate almost identical rankings on the E1, E2, mDice, and mHdis, which reflect that the functionality of the metric is equivalent to some extent. This may be the reason why the NICE I competition only adopted the E1 as the measure for ranking the methods. Further observations can see that the diverse results on the two methods. The method Lao Yang with the lowest sum rank 14 for E1 and E2 in iris segmentation is the best solution. Moreover, that method has achieved the best result on iris localization with a sum rank of 56. Even though our proposed method has achieved the second rank in overall comparison, our proposed method achieved better results on CASIA-Iris-M1 during evaluation. Using our collected dataset based on a mobile device might be why we have achieved better performance on the CASIA-Iris-M1 test set.

The segmentation and localization of the iris are important techniques in iris preprocessing when it comes to conventional iris identification, according to the authors. Therefore, in order to determine if a given entity is appropriate for usage as a preprocessing plug-in in an existing iris recognition system, a comprehensive evaluation of iris segmentation and localization performance must be conducted. As indicated in Table 2, the entries are arranged in descending order based on their rank sums on the iris segmentation and localization tasks. The rank-sum decreases in direct proportion to the overall performance. The research concludes by demonstrating that the suggested method provides excellent performance. We received the highest ranking on CASIA-Iris-M1 because the two-independent single-tasking model setup makes model training easier than the multi-tasking model used in the U-net approach. This is because our method uses a single single-tasking model setup that is less complicated than the U-net approach. We propose a new transfer learning training approach to enhance the model's generalization capabilities, as well as a model ensemble-based Test-Time Augmentation strategy to refine the prediction results, both of which are missing in our proposed method. As a last point, the adoption of MobileNetV2 as the backbone of our proposed approach results in a more robust feature extraction capacity than U-net.

*5.4. Qualitative Evaluation*

This part compares and contrasts several approaches in terms of anticipated binary iris segmentation masks and iris inner and outer boundaries. Two representative and challenging samples from the CASIA-Iris-Distance, CASIA-Iris-Complex-Occlusion, CASIA-Iris-Complex-Off-angle, CASIA-Iris-M1, and CASIA-Iris-Africa datasets were chosen separately to demonstrate the methods' iris segmentation and localization performance in different noncooperative environments, as shown in Fig 9, 10, and 11. The best of the applied methods' iris segmentation and localization results are shown in Fig 9, 10, and 11. They show that the best approach of CASIA-Iris-M1 is robust with various types of degraded iris pictures in noncooperative contexts but that further enhancements are needed to better cope with some difficult cases.

Table 2: Performance comparison between the proposed approach and existing methods; EyeCool, Lao Yang Sprint Team, and U-net (baseline) on iris localization and segmentation tasks. The best values are bolded based on their rank sum on all evaluation measures.

| Method | Iris Segmentation | | | Iris localization | | | Rank Sum |
|---|---|---|---|---|---|---|---|
| | E1 (Rank) | E2 (Rank) | Rank sum | mDice (Rank) | mHdis (Rank) | Rank Sum | |
| EyeCool | 34 | 34 | 68 | 30 | 28 | 58 | 112 |
| Lao Yang Sprint Team | 7 | 7 | 14 | 20 | 22 | 42 | 56 |
| U-net (baseline) | 92 | 92 | 184 | 106 | 115 | 221 | 405 |
| Proposed method | 19 | 19 | 38 | 21 | 23 | 44 | 82 |

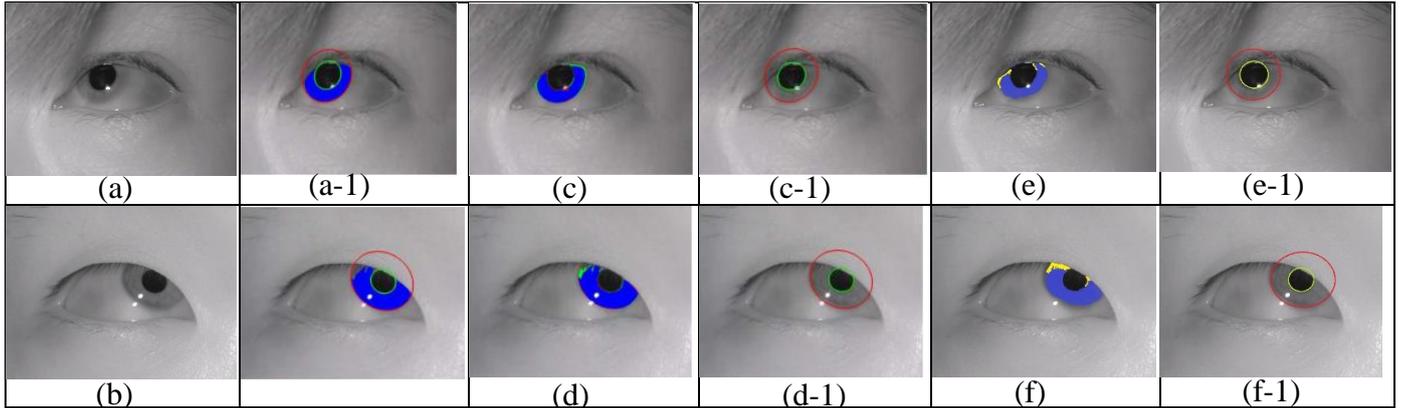

Fig 9: Samples of iris segmentation and localization results from best two methods (Two related work and our proposed method) on the CASIA-Iris-Complex-Off angle. a) real sample, a-1) grand truth for (a), b) real sample, b-1) grand truth for (b), c) Lao Yang Sprint Team's iris segmentation, c-1) Lao Yang Sprint Team's iris localization, e) our proposed method iris recognition, and e-1) our proposed method's iris localization. (d), (d-1), (f) and (f-1) results for methods like sample (a).

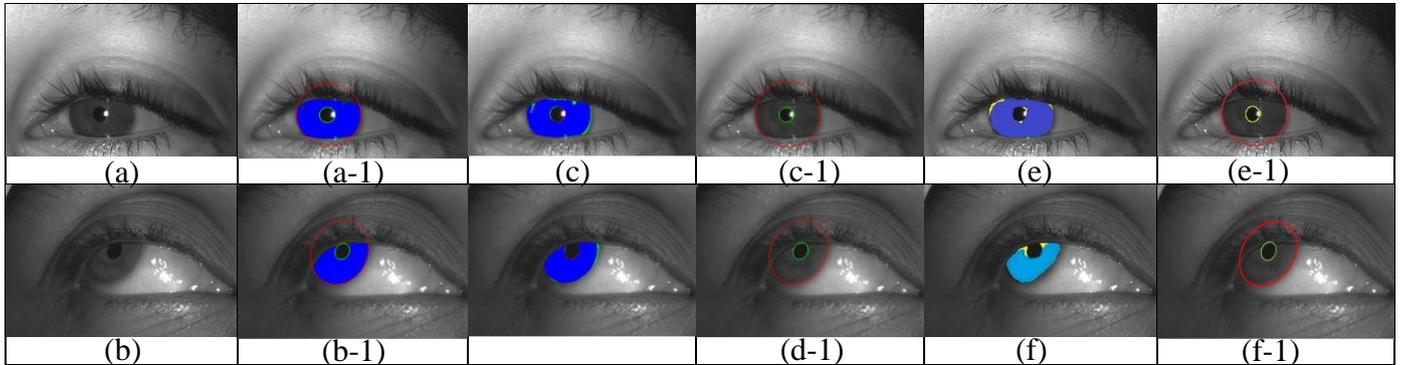

Fig 10: Samples of iris segmentation and localization result from the best two methods (Two related works and our proposed method) on the CASIA-Iris-Africa. a) real sample, a-1) grand truth for (a), b) real sample, b-1) grand truth for (b), c) Lao Yang Sprint Team's iris segmentation, c-1) Lao Yang Sprint Team's iris localization, e) our proposed method iris recognition, and e-1) our proposed method's iris localization. (d), (d-1), (f) and (f-1) results for methods like sample (a).

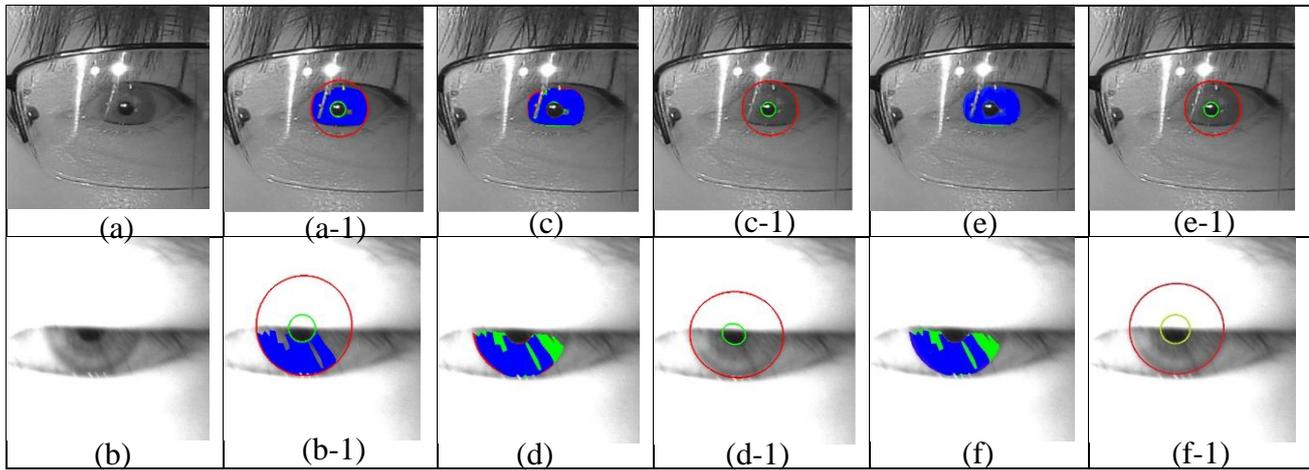

Fig 11: Samples of iris segmentation and localization result from the best two methods (Two related works and our proposed method) on the CASIA-Iris-M1. a) real sample, a-1) grand truth for (a), b) real sample, b-1) grand truth for (b), c) Lao Yang Sprint Team's iris segmentation, c-1) Lao Yang Sprint Team's iris localization, e) our proposed method iris recognition, and e-1) our proposed method's iris localization. (d), (d-1), (f) and (f-1) results for methods like sample (a).

## 6. Conclusions

This paper presents a U-Net method with a pre-trained MobileNetV2 deep neural network method for iris segmentation and localization. In this method, to extract better geometric constraints of inner and outer boundaries for both pupil and iris, we propose an encoder-decoder network to improve iris segmentation and localization performance. Meanwhile, we use three well-known publicly available NIR iris datasets, along with our collected dataset, which is also publicly available. We compare our proposed method with state-of-the-art methods on the three iris datasets that prove the top performance of our proposed method. As for future research, we will invest in enhancing the iris segmentation and localization efficiency to improve state-of-the-art NIR iris images datasets.

## Acknowledgments

The first author was supported by the European Union's Horizon 2020 research and innovation programmed under the Marie Skłodowska-Curie grant agreement No 675087.